\documentclass[sn-mathphys]{sn-jnl}

\jyear{2021}%

\theoremstyle{thmstyleone}%
\theoremstyle{thmstyletwo}%

\theoremstyle{thmstylethree}%

\usepackage{natbib}
\usepackage{url}
\usepackage{lineno,hyperref}

\begin{document}
	
\title[Article Title]{A Reinforcement Learning-based Offensive semantics Censorship System for Chatbots}

\author[1]{\fnm{Shaokang} \sur{Cai}}\email{a865316182@gmail.com}
	
\author*[1]{\fnm{Dezhi} \sur{Han}}\email{dzhan@shmtu.edu.cn}
	\equalcont{These authors contributed equally to this work.}
	
\author*[1,3]{\fnm{Dun} \sur{Li}}\email{lidunshmtu@outlook.com}
\equalcont{These authors contributed equally to this work.}

\author[2]{\fnm{Zibin} \sur{Zheng}}\email{zhzibin@mail.sysu.edu.cn}

\author[3]{\fnm{Noel}\sur{Crespi}}\email{noel.crespi@mines-telecom.fr}
\equalcont{These authors contributed equally to this work.}
	
\affil*[1]{\orgdiv{College of Information Engineering},
\orgname{Shanghai Maritime University}, \orgaddress{
\city{Shanghai}, 
\postcode{201306}, \country{China}}}
	
\affil[2]{\orgdiv{School of Software Engineering}, \orgname{Sun Yat-sen University}, \orgaddress{
\city{Zhuhai}, 
\postcode{519082}, 
\country{China}}}
	
\affil[3]{\orgdiv{Telecom SudParis}, \orgname{IMT, Institut Polytechnique de Paris}, \orgaddress{
\city{Paris}, \postcode{91000}, \country{France}}}
	
	
	
\abstract{The rapid development of artificial intelligence (AI) technology has enabled large-scale AI applications to land in the market and practice. However, while AI technology has brought many conveniences to people in the productization process, it has also exposed many security issues. 
Especially, attacks against online learning vulnerabilities of chatbots occur frequently. 
Therefore, this paper proposes a semantics censorship chatbot system based on reinforcement learning, which is mainly composed of two parts: the Offensive semantics censorship model and the semantics purification model. 
Offensive semantics review can combine the context of user input sentences to detect the rapid evolution of Offensive semantics and respond to Offensive semantics responses. 
The semantics purification model For the case of chatting robot models, it has been contaminated by large numbers of offensive semantics, by strengthening the offensive reply learned by the learning algorithm, rather than rolling back to the early versions.
In addition, by integrating a once-through learning approach, the speed of semantics purification is accelerated while reducing the impact on the quality of replies. 
The experimental results show that our proposed approach reduces the probability of the chat model generating offensive replies and that the integration of the few-shot learning algorithm improves the training speed rapidly while effectively slowing down the decline in BLEU values.}

	
	
	
\maketitle
	
\section{Introduction}\label{sec1}
As human-machine interaction technology continues to advance, the development of information technology, represented by Internet technology, has made dialogue-based interaction technology more and more important and widely used. 
People use the Internet to access a large amount of information that is relevant to their lives and work, and language is one of the most direct types of information, so it is particularly important to get the right and important information back to us from the many linguistic messages available. 
Artificial intelligence (AI), often thought of as computer systems with human-like thinking and capabilities \cite{kok2009artificial}\cite{poole2010artificial}, is used in a wide range of applications such as voice chat, autonomous driving, social media, gaming, industry, and even replacing humans in tedious, repetitive tasks \cite{li2022blockchain,li2022mfvt,li2019panoramic,cai2022hybrid,zhang2019transmission}. 

Specifically, chatbots have been widely used in business and government affairs.
Chatbots are computer programs that can fully interact with users using natural language based on the input \cite{adamopoulou2020overview}\cite{khan2018introduction}. 
Compared to traditional search engines, chatbots can extract the information the user needs from the vast amount of information available, but with a greater emphasis on the stickiness of the interaction with the user i.e. they do not want the user to leave as soon as possible and therefore have a better interaction effect \cite{li2022moocschain}. 	
Nowadays, with the popularity of various smartphones, many enterprises have invested huge manpower and material resources in the technical exploration and product landing of chatbots and achieved good results, such as Microsoft's chatbot Xiaobing, Apple's personal voice assistant Siri, etc., is very excellent and practical chatbot products.

However, the online learning technology of chatbots, which allows them to learn and develop as they interact with users, constantly enriches the diversity of the response corpus while also making them subject to some influences related to the user's language use in the learning process \cite{hill2015real}.
A hacker or offensive user can use the online learning interface of a chatbot to teach extreme semantics to the robot, resulting in an improper semantics by the chatbot, violating local laws and regulations \cite{park2021use, li2021design}. 
For example, only a few hours after Tay is online, offensive users exploit its training vulnerability to teach Tay racist semantics (including racial discrimination, gender discrimination, propaganda of violence, white supremacy, and genocide), resulting in the offline of the product.

So far, the key method for preventing the online learning process of chatbots from being contaminated is offensive language response detection, also known as semantics censorship.
However, the datasets (such as YouTube-based movie reviews\cite{dadvar2013improving} and Twitter-based offensive language response datasets\cite{xiang2012detecting}) used in current chatbot research have the disadvantage of focusing only on a single offensive response sentence and ignoring user input.
This is because even the same response sentences in different contexts can have different classification results when faced with different input sentences. 
The user input sentence is the key to the semantics review of the reply sentence of the chatbot. 
However, the existing work does not take this into account.

To fill this gap, we propose a semantics censorship chatbot system (RLC) based on reinforcement learning, which is mainly composed of two parts: Offensive semantics censorship model and semantics purification model, aiming at the current situation of chatbots in which users spread a large number of offensive languages in the network, which affects continuous online learning,
Offensive semantics review can combine the context of user input sentences to detect the rapid evolution of Offensive semantics and respond to Offensive semantics responses. 
The semantics cleansing model is designed for situations where the chatbot model has been contaminated with large amounts of offensive semantics, and through reinforcement learning algorithms can "forget" learned offensive replies rather than roll back to earlier versions.

Specifically, the main contributions of this study are as follows.

\begin{itemize}
	\item{In this paper, we propose an offensive semantics censorship model based on a bi-directional gated recurrent unit network (Bi-GRU) of attention mechanisms forming an encoder-decoder structure. The encoder encodes the user input sentence into a context vector and later embeds the context vector into each time step of the reply sentence classification. This model architecture is used to better fit the task of semantics censorship for chatbots.}
	\item {We propose a reinforcement learning-based semantics purification algorithm. The algorithm can forget learned offensive replies when the chatbot model has been contaminated by reinforcement learning methods, rather than rolling back to some earlier version. Experiments on the offensive replies dataset demonstrate the ability to reduce the probability of chat models generating offensive replies by this algorithm. }
	\item {This paper incorporates a few-shot learning approach into the semantics purification algorithm, allowing the algorithm to perform semantics purification quickly while minimizing forgetting previously learned basic syntax. Experiments on the Offensive reply dataset demonstrate that the integration of the less-sample learning algorithm improves training speed while reducing the impact on reply syntax.}
\end{itemize}

The rest of this paper is arranged as follows. 
Section \ref{sec2} gives a brief overview of the related work. 
Section \ref{sec3} introduces the specific steps of the Offensive semantics censorship model and semantics purification algorithm. 
In Section \ref{sec4}, we first introduce the environment and parameter setting of this experiment and analyze the comparison between our proposed model and the existing most advanced model. Finally, section \ref{sec5} summarizes the full text and looks forward to the future research direction.
	
\section{Related Work}\label{sec2}

In this section, we survey the researches related to the censorship of chatbots.
We present the training methods for chatbot models in section \ref{sec2.1} and the reinforcement learning-based models for offensive semantics detection in section \ref{sec2.2}.

\subsection{Training methods for chatbot models}\label{sec2.1}
Online Learning (OL) is a training method for machine learning models that can be updated in real-time and quickly based on online feedback data so that they can reflect changes online promptly. li \cite{li2016dialogue} constructs a simulation environment in a reinforcement learning framework to improve the BOT conversational actions based on different types of feedback signals from the teacher (conversation partner) on the chatbot's ability to respond. The digital feedback was passed to the chatbot through a reinforcement learning approach, allowing the authors to process textual feedback using forward prediction methods. David \cite{abel2017agent} proposes to improve the learning performance of the chatbot by incorporating human feedback into a neural dialogue model through online learning, and thus online interaction with humans. Asghar \textit{et al.} \cite{asghar2016deep} proposes offline two-stage supervised learning and online Human in the loop (HIL) active learning for dialogue generation. The model interacts with real users and gradually learns from their feedback in each round of conversation, with different feedback affecting the chatbot's predicted response to different prompts. However, the above models share a common flaw: people may use these fast and unrestricted learning capabilities to teach online learning chatbots to produce Offensive responses.

\subsection{Reinforcement learning-based model for offensive semantics detection}\label{sec2.2}
Offensive semantics review can be attributed to either text classification or sentiment analysis \cite{du2017convolutional}\cite{yang2016hierarchical}\cite{li2015tuning}\cite{liu2016recurrent}.
Ravi \cite{ravi2015survey} and Enas \cite{Khalil2020DeepLA} provide a review of deep learning algorithms in sentiment analysis. Specifically, for the offensive semantics review task, AlLouch \cite{Allouch2019DetectingST} constructed a dataset of sentences that could be harmful to children's thinking and proposed a voting method for detection using multiple classifiers.
Razavi \cite{Razavi2010OffensiveLD} proposed a multilayer Bayesian Offensive classifier that performs feature detection on Offensive semantics at three different conceptual levels with good results \cite{Spertus1997SmokeyAR}\cite{li2015tuning}\cite{liu2016recurrent}.
Chkroun \cite{Chkroun2018SafebotAS} proposed a secure collaborative chatbot called Safebot. First, Safebot detects users posting offensive semantics and marks them as offensive users by using a offensive semantics detection model. 
The responses entered by the offensive user are then stored in a offensive dataset. During the "learning state" Safebot searches the offensive dataset to determine which response is closest to the response entered by the user. If the input response is determined to be the closest to an entry in the offensive dataset, Safebot blocks the learning of the user's input response and warns the user.

\section{System Model and Design}\label{sec3}
In this section, we introduce the offensive semantics review algorithm and semantics purification algorithm of pre-knowledge and RLC. 
Section \ref{sec3.1} first provides definitions of different offensive semantics. 
In Section \ref{sec3.2}, we introduce our semantics review algorithm. 
Finally, Section \ref{sec3.3} introduces the semantics purification algorithm using reinforcement learning.

\subsection{Offensive semantics}\label{sec3.1}
To clearly analyze the data set, we created the following category according to the response: offensive semantics, danger semantics, and non-incompatible semantics.

\textbf{Violent semantics}: Textual surfaces in the response sentences contain aggressive words. This type of semantics can be detected simply by keyword or rule-based methods.
 
\textbf{Dangerous semantics}: Response sentences in which the textual surface does not contain aggressive words, but the semantics contains the meaning of aggression. This category can be detected by semantic-based machine learning methods for response utterances. 

\textbf{Offensive semantics}: the response sentence does not contain either of the above, but has a violation meaning when combined with the context of the input sentence. For example, the same response "He is a great man", in response to the questions "What do you think of Newton" and "What do you think of Bin Laden? The meanings expressed in the responses to the questions "What do you think of Newton" and "What do you think of Bin Laden" are different. (Note: In this case, when the input sentence is changed, the reply sentence may become the normal reply.)

\subsection{semantics censorship algorithms}\label{sec3.2}
\begin{figure}[h]%
	\centering
	\includegraphics[width=0.9\textwidth]{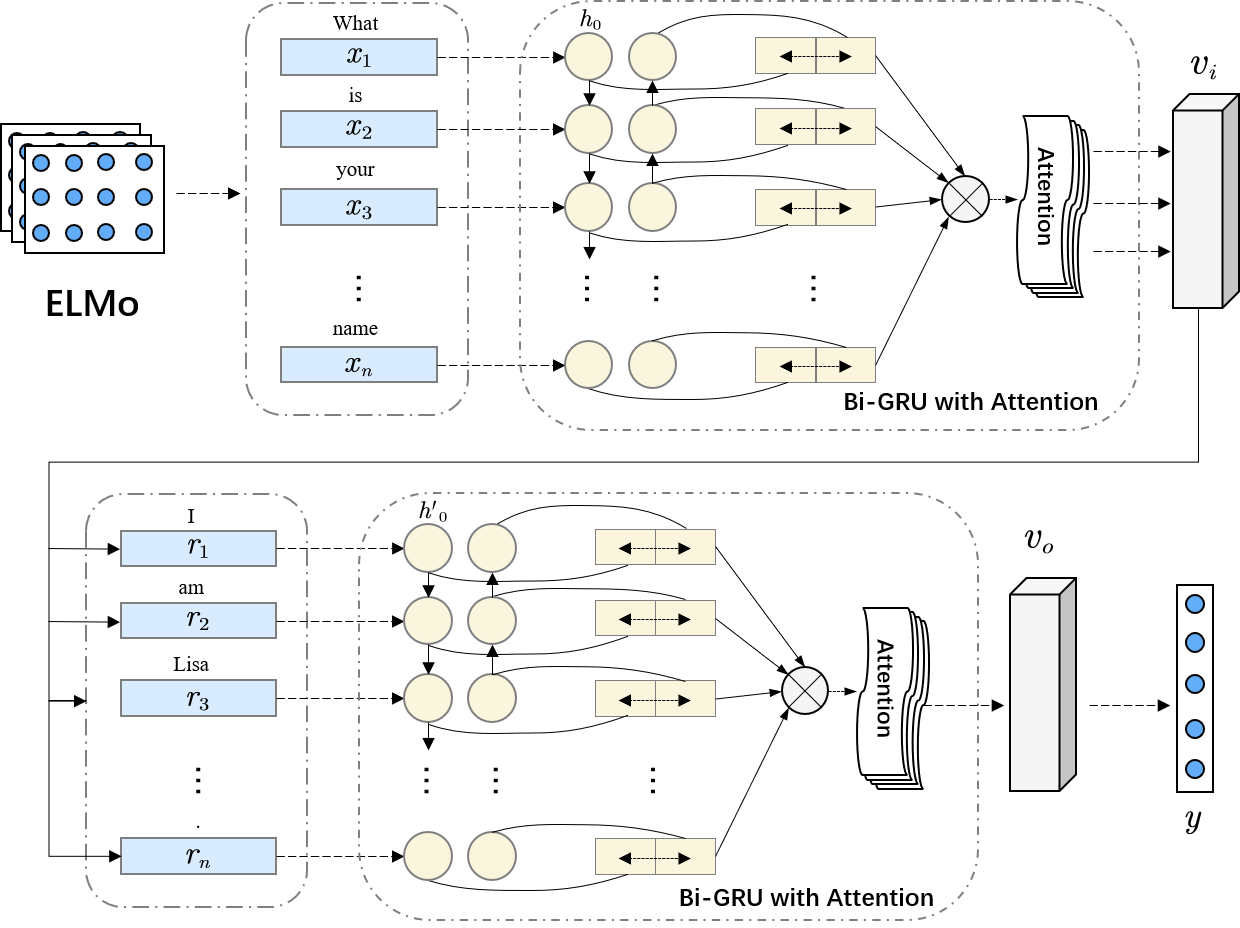}
	\caption{Schematic diagram of the general framework of the model, where $x$ is the input and $r$ is the response. $h$ is the implied state of the input, $h^{\prime}$ is the implied state of the response, $v_{i}$ and $v_{o}$ are vectors summarising the information in the input/output sentences, and $y$ is the output of the model.}\label{fig1}
\end{figure}

Directly connecting inputs and responses to classifiers enhances the long correlation problem of RNN-based models \cite{Hochreiter1997LongSM}. Therefore, we propose a hybrid model Bi-GRU paired with an attention mechanism to censor the user's responses to offensive semantics. The structure of the model is shown in Fig. \ref{fig1}. 
The model mainly consists of an embedding layer, an encoding layer, and a decoding layer, where the embedding layer is responsible for converting characters into vectors and the encoding layer partially encodes the user's input into a character vector representing the semantics of the input context. 

\subsubsection{Embedding layer}\label{subsubsec3.21}
\begin{figure}[h]%
	\centering
	\includegraphics[width=0.9\textwidth]{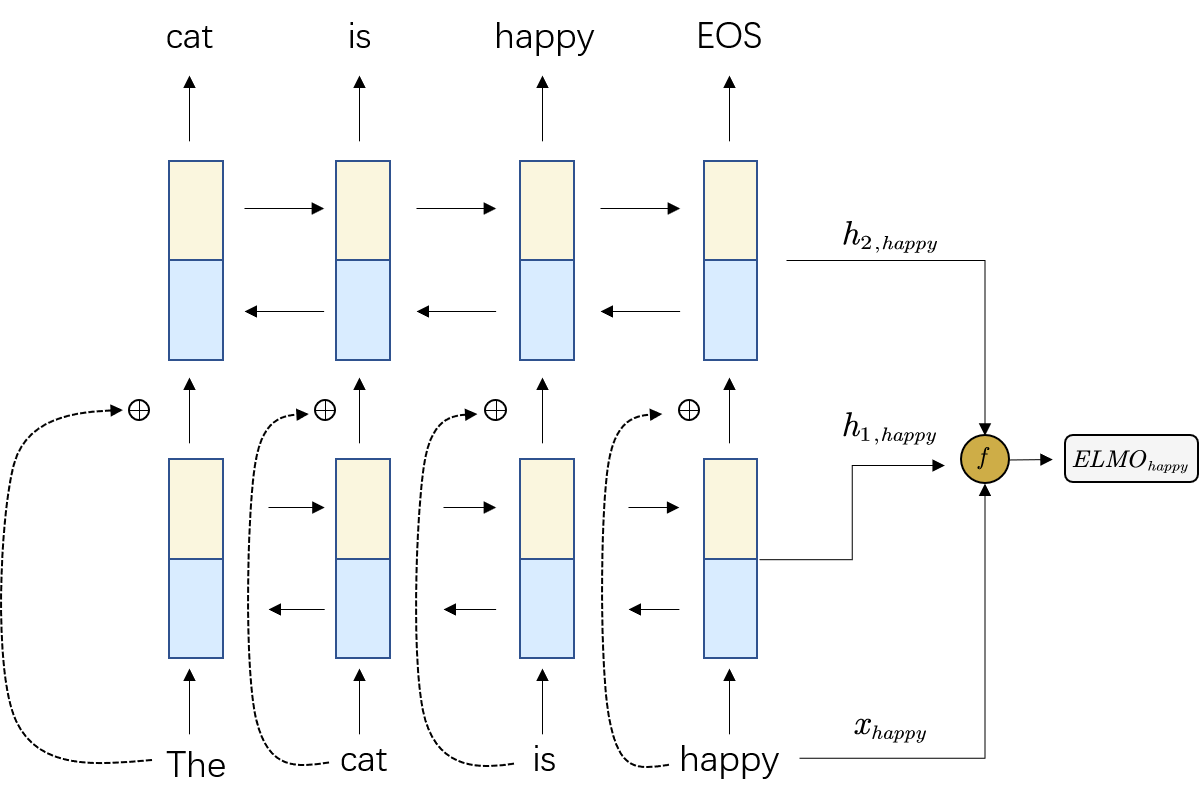}
	\caption{ELMo's specific representation of 'happy'.}\label{fig2}
\end{figure}

Word embedding is a distribution-based idea: semantic (or morphological) related words often appear in a similar context. By a continuous low dimensional vector, each word is used to effectively retain the semantic information of the term. 
This paper uses pre-training ELMO (Embeddings from Language Models) as characters embedded \cite{Peters2018DeepCW}. 
ELMO is more advantageous compared to other traditional embedded (such as Glove and Word2Vec) because it encapsulates the context in the word characteristic representation. ELMO uses a two-dimensional LSTM to learn words and their context, which enables ELMO to learn more-related words related to contexts in higher dimensions and learn syntax knowledge in lower dimensions. 
Fig \ref{fig2} shows an example of how to generate ELMO through a binding bidirectional hidden characterization.

\subsubsection{Encoding layer}\label{subsubsec3.22}
Recurrent neural network (RNN) units cannot memorize values for long periods \cite{park2021use}.
To solve the gradient vanishing problem in RNNs, researchers have proposed gated recursive units (GRU) and long short-term memory (LSTM), respectively, to replace hidden layer neurons with memory units that store early sequence data \cite{Sherstinsky2018FundamentalsOR}. 
Since the user responses have shorter sequences, the choice of GRU reduces the training time without loss of accuracy.

This paper employed a two-way GRU (BI-GRU) as an encoder in the coding layer, which receives the input sequence and encapsulates the information into internal state vectors.  
The GRU network has two gate structures, update gate $z_{t}$ and reset gate $r_{t}$, $z_{t}$ is used to indicate the reception of information by the cell in the previous time step, with higher values indicating that more information from the previous time step is remembered. $r_{t}$ is used to indicate the extent to which information from the previous time step is ignored, with a smaller value indicating that more information is forgotten. 
At a given point in time, the hidden state of the GRU is calculated as shown in Eq.\ref{eq1} - \ref{eq4}.

\begin{equation}
	\label{eq1}
 \begin{aligned}
	r_{t}=\sigma\left(W_{r} x_{t}+U_{r} h_{t-1}+b_{r}\right)
 \end{aligned}
\end{equation}

\begin{equation}
	\label{eq2}
	\begin{aligned}
	 z_{t}=\sigma\left(W_{z} x_{t}+U_{z} h_{t-1}+b_{z}\right)
	\end{aligned}
\end{equation}

\begin{equation}
	\label{eq3}
	\begin{aligned}
		\tilde{h}_{t}=\tanh \left(W_{h} x_{t}+U_{h}\left(r_{t} \odot h_{t-1}\right)+b_{h}\right)
	\end{aligned}
\end{equation}

\begin{equation}
	\label{eq4}
	\begin{aligned}
		h_{t}=\left(1-z_{t}\right) \odot h_{t-1}+z_{t} \odot \widetilde{h_{t}}
	\end{aligned}
\end{equation}

where $r_{t}$ denotes the update gate, $z_{t}$ denotes the reset gate, $h_{t-1}$ denotes the last moment hidden state, $\odot$ denotes the element multiplication, z denotes the input sequence information, $W$ and $U$ are the weight matrices, and $\sigma$ is the sigmoid function.

Combining the forward and backward hidden layers gives the output of the bi-directional GRU encoder $h_{t}=\left[\vec{h}_{t}, \overleftarrow{h}_{t}\right]$, combined with the forward hidden layer $\vec{h}_{t}=\left(\vec{h}_{1}, \vec{h}_{2} \cdots \vec{h}_{n}\right)$ and the backward hidden layer $\overleftarrow{h_{t}}=\left(\overleftarrow{h_{1}}, \overleftarrow{h_{2}} \cdots \overleftarrow{h_{n}}\right)$, where $n$ is the length of the sentence.
Thus, in contrast to the unidirectional GRU, the Bi-GRU allows the capture of information from the previous and the next point in time to make predictions about the current state. In contrast to unidirectional GRU, Bi-GRU can understand the meaning and context of the sentences. We add an attention layer after the bi-directional GRU encoder. 
The attention layer learns the weight of each word and increases the weight share of important features as Eq. \ref{eq5} - \ref{eq7}.

\begin{equation}
	\label{eq5}
	\begin{aligned}
		u_{t}=\tanh \left(W_{u} h_{t}+b_{u}\right)
	\end{aligned}
\end{equation}

\begin{equation}
	\label{eq6}
	\begin{aligned}
		\alpha_{t}=\frac{\exp \left(u_{t} u_{a}\right)}{\sum_{t} \exp \left(u_{t} u_{a}\right)}
	\end{aligned}
\end{equation}

\begin{equation}
	\label{eq7}
	\begin{aligned}
    	v=\sum_{t} \alpha_{t} h_{t}
	\end{aligned}
\end{equation}

where $u_{t}$ is a non-linear transformation of $h_{t}$, 
$u_{a}$ represents the context vector, which is randomly initialized and learned jointly with other parameters as the training process progresses, 
$v$ is a vector containing the semantics of the input sentence, $\alpha_{t}$ is the attention weight, and each word in the input sentence is given an attention weight $\alpha$. 
The value of the weight $\alpha$ is restricted between 0 and 1 and determines which implicit states $h$ in the input sentence have a higher weight.

\subsubsection{Decoding layer}\label{subsubsec3.23}
In the encoding layer, Bi-GRU encodes the user input sentence into a vector $v \in \mathbb{R}^{n \times 1}$ that represents the semantics of the input sentence, where $n$ denotes the length of the sentence. 
This semantic vector is then embedded in each time step of the reply sentence classifier. 
The GRU conversion formula for the encoder part is as Eq. \ref{eq8}.

\begin{equation}
	\label{eq8}
	\begin{aligned}
	 h_{t}=G R U\left(h_{t-1}, x_{t}\right)
	\end{aligned}
\end{equation}

where $h_{t}$ is the output of the time step $t$, 
$x_{t}$ is the input on the time step $t$, and $h_{t-1}$ is the hidden state of the time step $t-1$. GRU is shorthand for the transformation equation.

In the GRU transformation formula for the decoding layer, we combine the hidden state of the previous time step $h_{t}$, the word vector $x_{t}$ in the current time step, and the input semantic vector $v$, as Eq. \ref{eq9} - \ref{eq12}.

\begin{equation}
	\label{eq9}
	\begin{aligned}
		r_{t}=\sigma\left(W_{r}\left[x_{t}, h_{t-1}, v\right]+b_{r}\right)
	\end{aligned}
\end{equation}

\begin{equation}
	\label{eq10}
	\begin{aligned}
		z_{t}=\sigma\left(W_{z}\left[x_{t}, h_{t-1}, v\right]+b_{z}\right)
	\end{aligned}
\end{equation}

\begin{equation}
	\label{eq11}
	\begin{aligned}
		\widetilde{h_{t}}=\tanh \left(W_{h}\left[x_{t}, h_{t-1}, v\right]+b_{h}\right)
	\end{aligned}
\end{equation}

\begin{equation}
	\label{eq12}
	\begin{aligned}
		h_{t}=\left(1-z_{t}\right) \odot h_{t-1}+z_{t} \odot \widetilde{h}_{t}
	\end{aligned}
\end{equation}

With this method, the semantic vector of the input sentence is then embedded in each time step of the decoder GRU hidden layer and used to help predict the security score $\hat{y}$ of the reply sentence as Eq. \ref{eq13}.

\begin{equation}
	\label{eq13}
	\begin{aligned}
	\hat{y}=\tanh \left(W_{y} v_{y}+b_{y}\right)
	\end{aligned}
\end{equation}

Where $v_{y}$ is the semantic vector of the output sentence and $\hat{y}$ ranges from $[-1,1]$.
At this point, the optimization objective function of the model is obtained as Eq. \ref{eq14}.

\begin{equation}
	\label{eq14}
	\begin{aligned}
    	L=-\left(\frac{1}{2}(1+y) \ln \hat{y}+\frac{1}{2}(1-y) \ln (1-\hat{y})\right)
	\end{aligned}
\end{equation}

Where $y$ is the true label taking values of -1 and 1. 
Since $\hat{y}$ takes values in the range $[-1,1]$, the above equation has a slightly different form than the cross-entropy loss when the labels are 0 and 1.

The global flow of the semantics censorship algorithm is as follows: 
Firstly, the chatbot generates a set $R_{c}$ of $k$ candidate responses based on the input sentence $s$. 
Then, the security score $\hat{y}$ is calculated based on Eq. \ref{eq14}.
If the security score is greater than 0, the score and responses are added to the temporary response set $temp$. 
Finally, the temporary response set is sorted and the security response set $r$ with the highest score is filtered.

\subsection{semantics purification algorithm}\label{sec3.3}

\begin{figure}[h]%
	\centering
	\includegraphics[width=0.9\textwidth]{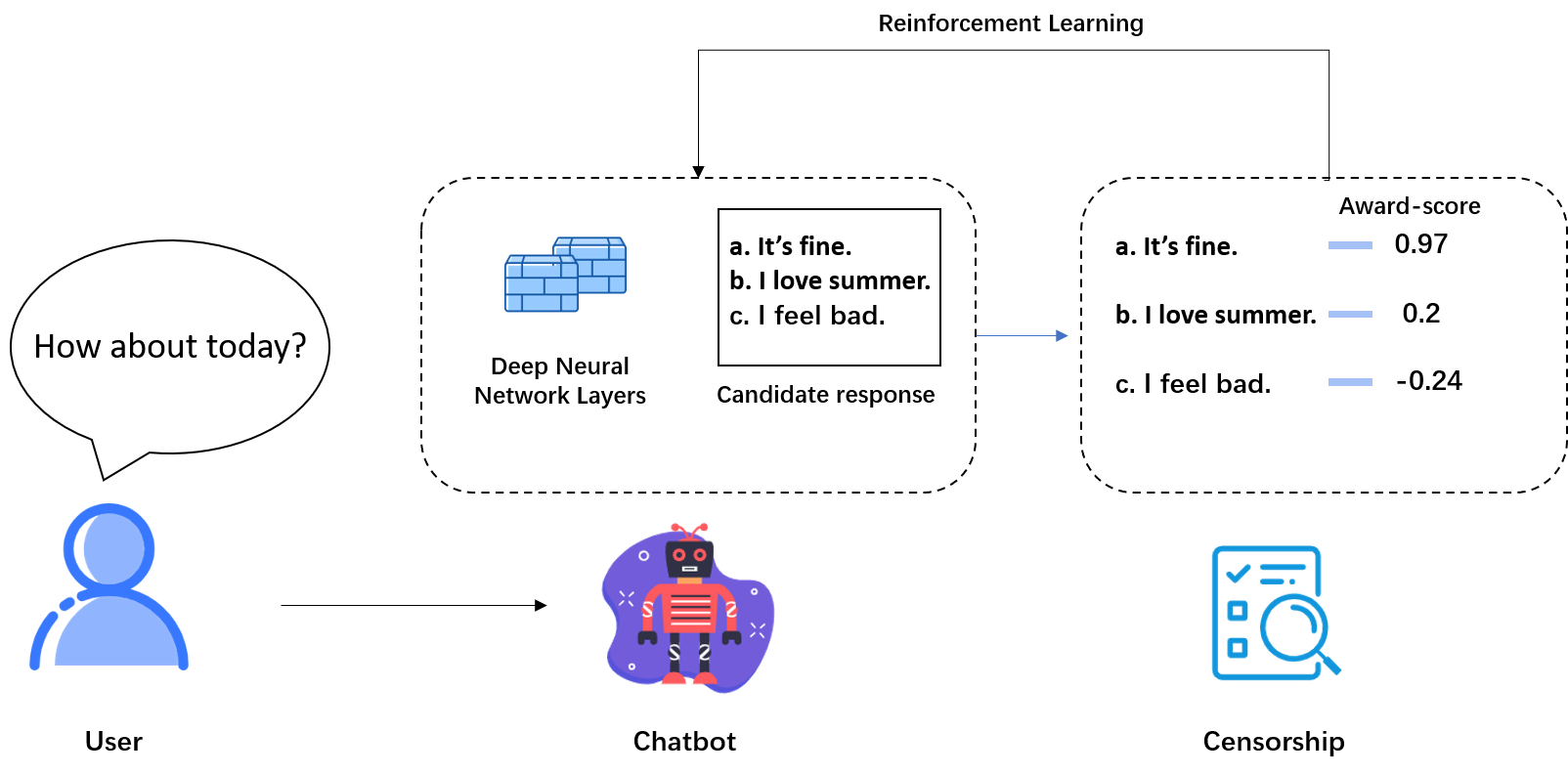}
	\caption{Flow diagram of the semantics purification algorithm.}\label{fig3}
\end{figure}

Due to the uncontrolled and unrestricted online learning of chatbots, malicious users can interfere with the learning algorithms of chatbots through large batches of offensive or insulting comments, causing them to generate invasive responses when conversing with other normal users, causing property and psychological damage to companies and users alike. Therefore, we purify the polluted chatbots through a reinforcement learning approach. The flow of the semantics purification algorithm is shown in Fig \ref{fig3}. 
In our semantics purification algorithm, the chatbot accepts user input sentences and outputs $k$ candidate responses. 
The input sentences and candidate responses are then sent together to the semantics review model, which will generate a return value (i.e. a safety score) for each candidate response, which will be fed back to the chatbot as a reward function for reinforcement learning. 
Through the reinforcement learning process, the model will reduce the probability of producing offensive responses. In addition, the Few-shot Learning method is introduced to reduce the amount of input to the replies so that the quality of the replies generated can be influenced as little as possible in the semantics cleaning process.

\subsubsection{Reinforcement learning with reward functions}\label{subsubsec3.31}
Reinforcement Learning (RL) is a learning method that makes serialized decisions based on feedback from a given environment in the hope of maximizing returns. In reinforcement learning, we refer to the model to be learned as the robot (Agent). 
The robot selects its action to be performed by observing the State in the Environment and receives an award based on the action performed by the robot in the current state. There are various augmentation learning algorithms, which can be divided into Policy-based RL and Value-based RL approaches. 
The utterance generation model of chatbots is a sequence-to-sequence model, and the sequence-to-sequence model has a large action space (i.e. each word corresponds to an action), so if we use the value-based approach we need to provide a value estimate for each word in the vocabulary, so we use the policy-based distribution approach to feed the chatbot.
The strategy gradient directly outputs the probability distribution for each action at the next moment based on the observed environment. The formalization is defined as Eq. \ref{eq15}.

\begin{equation}
	\label{eq15}
	\begin{aligned}
	a_{t} \sim p\left(a_{t} \mid a_{1: t-1}, s_{t-1}\right)
	\end{aligned}
\end{equation}

where $a_{1: t-1}$ is the sequence of actions taken in the past moment and $s_{t-1}$ is the moment state of $t-1$.
The action probability distribution is the policy distribution, denoted as $\pi_{\theta}\left(a_{t} \mid a_{1: t-1}, s_{t-1}\right)$, and $\theta$ is the parameter to be optimized.
The response words $y_{t}$ predicted by the model at moment $t$ can be seen as action $a_{t}$, the action space is the size of the vocabulary, and the input sentence $x$ can be seen as state $s$.
The formal definition of Eq. \ref{eq15} can be modified in the utterance generation model for chatbots as Eq. \ref{eq16}.

\begin{equation}
	\label{eq16}
	\begin{aligned}
	y_{\mathrm{t}} \sim p\left(y_{\mathrm{t}} \mid y_{1: \mathrm{t}-1}, x\right)
	\end{aligned}
\end{equation}

The chatbot's utterance generation model (model parameter t) generates a reply sentence $y$ when it receives a user input sentence $x$.
The semantics censorship algorithm takes $x$ and $y$ as input to obtain a payoff value r, which represents whether the reply is offensive or not, and the payoff value ranges from $[-1,1]$. The goal of the reinforcement learning algorithm is to maximise the desired return value obtained.
Where the expected payoff value is as Eq. \ref{eq17}.

\begin{equation}
	\label{eq17}
	\begin{aligned}
	\bar{A}_{\theta}=\sum_{x} p(x) \sum_{y} A(x, y) p_{\theta}(y \mid x)
	\end{aligned}
\end{equation}

where $p(x)$ is the probability of occurrence of the input sentence $x$, $p_{\theta}(y \mid x)$ is the probability that the chat model with parameter $\theta$ will reply with sentence $y$ when the input sentence is $x$, and $A(x, y)$ is the payoff function.
The semantics purification algorithm uses the semantics censorship model as the payoff function, with the final payoff value being the output of the semantics censorship algorithm as Eq. \ref{eq18}.

\begin{equation}
	\label{eq18}
	\begin{aligned}
		A(x, y)=\hat{y}=\tanh \left(W_{y} v_{y}+b_{y}\right)
	\end{aligned}
\end{equation}

Where the return value is between $[-1,1]$, the return value obtained when the reply sentence is an offensive reply is a negative number. 
Conversely, the return value obtained when the reply sentence is normal is a positive number.
The training phase maximizes the desired payoff value by updating the parameters $\theta$ of the chatbot model as Eq. \ref{eq19}.

\begin{equation}
	\label{eq19}
	\begin{aligned}
		\theta^{*}=\underset{\theta}{\operatorname{argmax}} \bar{A}_{\theta}
	\end{aligned}
\end{equation}

where the function $\arg \max _{\theta} A$ denotes finding a value $\theta$ such that $A$ obtains its maximum value. 
The parameters are updated by Eq. \ref{eq20}.

\begin{equation}
	\label{eq20}
	\begin{aligned}
\theta=\theta+\alpha \nabla \bar{A}_{\theta}
	\end{aligned}
\end{equation}

where $\nabla \bar{A}_{\theta}$ is the gradient of return value expectation and $\alpha$ is the learning rate. 
Specifically, $\nabla \bar{A}_{\theta}$ is calculated as Eq. \ref{eq21}.

\begin{equation}
	\label{eq21}
	\begin{aligned}
		\begin{aligned}
			& \nabla \bar{A}_{\theta}=\sum_{x} p(x) \sum_{y} A(x, y) \nabla p_{\theta}(y \mid x) \\
			=& \sum_{x} p(x) \sum_{y} A(x, y) p_{\theta}(y \mid x) \frac{\nabla p_{\theta}(y \mid x)}{p_{\theta}(y \mid x)} \\
			=& \sum_{x} p(x) \sum_{y} A(x, y) p_{\theta}(y \mid x) \nabla \log p_{\theta}(y \mid x) \\
			=& E_{x \sim p(x), y \sim p_{\theta}(y \mid x)}\left[A(x, y) \nabla \log p_{\theta}(y \mid x)\right.
		\end{aligned}
	\end{aligned}
\end{equation}

In practice, a random sample of $N$ data is used to approximate the expected value as the true probability distribution cannot be calculated for large-scale data. 
In addition, to alleviate the high variance problem of the model, the value of the return function is subtracted from the baseline value $t$ as Eq. \ref{eq22}.

\begin{equation}
	\label{eq22}
	\begin{aligned}
		\begin{aligned}
\nabla \bar{A}_{\theta} \approx \frac{1}{N} \sum_{i=1}^{N}\left(A\left(x^{i}, y^{i}\right)-t\right) \nabla \log p_{\theta}\left(y^{i} \mid x^{i}\right)
		\end{aligned}
	\end{aligned}
\end{equation}

where $N$ is the number of random samples and the baseline value $t$ is the mean value of the observed return values.
Finally, the objective function for reinforcement learning in the semantics purification algorithm can be obtained as Eq. \ref{eq23}.

\begin{equation}
	\label{eq23}
	\begin{aligned}
	J(\theta)=\frac{1}{N} \sum_{i=1}^{N}\left(A\left(x^{i}, y^{i}\right)-b\right) \log p_{\theta}\left(y^{i} \mid x^{i}\right)
	\end{aligned}
\end{equation}

The flow of the semantics purification algorithm is shown in the algorithm, where the set of user input sentences and the currently contaminated model $M_{\theta}$ is input and the purified chatbot utterance generating model $M_{\theta}$ is output.
Each input sentence is first iterated through and fed into the chat model.
The chat model in line 2 samples the input sentences to generate $K$ responses.
The semantics review algorithm then calculates the return value based on each of the $K$ generated replies and the corresponding input sentences.
Finally, it determines whether any of the generated candidate responses have a safety score (payoff value) less than 0. 
If so, the policy gradient is used to update the model parameters.

\subsubsection{semantics purification algorithms based on few-shot learning}\label{subsubsec3.32}

Less sample learning is a method of transfer learning, which aims to learn information from a small number of training samples. 
We use a small amount of semantics review model feedback to clean up a small amount of contaminated chatbot model to reduce the probability of generating aggressive recovery. Increasing the learning rate is the most effective and convenient method to achieve less sample learning. But if the learning rate is too high, it can lead to reinforcement learning destroying the basic syntax already learned. The semantics purification algorithm needs fast semantics purification while avoiding the impact of the quality of the reply sentence, so simply improving the learning rate is not suitable for this algorithm. 

This paper only rewards (or penalizes) the first candidate's reply to quickly  select a normal reply. Since only the first candidate response is affected by the loss function, it has little effect on the candidate response generated later in the reinforcement learning process. The final objective function is as Eq. \ref{eq24}.

\begin{equation}
	\label{eq24}
	\begin{aligned}
J(\theta) \approx \frac{1}{N} \sum_{i=1}^{N}\left(A\left(x^{i}, y^{i}\right)-t\right) \sum_{j=1}^{n} \operatorname{safe}\{\cdot\} \log P_{\theta}\left(y_{j}^{i} \mid x^{i}, y_{1}^{i} \ldots y_{j-1}^{i}\right)
	\end{aligned}
\end{equation}

where $n$ is the length of the reply, $N$ is the number of random samples, and is the secure reply function: 
For secure reply functions: $safe\{normal\}=1$, otherwise $safe\{offensive\}=0$.

\section{Experiments and Performance Analysis}\label{sec4}
In this section, we experimentally evaluate the model proposed in this paper. 
The experiments verify two main aspects.
\begin{itemize}
	\item{The extent to which the semantics purification algorithm reduces the probability of generating aggressive replies from chatbots.}
	\item {The effect of introducing few-shot learning on the speed of training convergence and the quality of reply sentence generation.}
\end{itemize}

\subsection{Experimental preset}\label{sec4.1}
\subsubsection{Experimental environment configuration}\label{sec4.1.1}
All experiments in this paper are conducted on a cloud server with 12 CPU cores and a P4000 GPU. all code was developed on Python 3, based on the Pytorch 1.7.1 deep learning framework. Details of the equipment used to run the experiments are shown in Tab. \ref{tab1}.

\begin{table}[ht]
	\centering
	\caption{Hardware and software resources.}
	\label{tab1}
	\begin{tabular}{c|c}
		\toprule
		\hline
\textbf{Hardware}  &  \textbf{Configuration}        \\
	\hline
CPU    & Intel(R) Xeon(R) Silver 4116 CPU @ 2.10GHz \\
GPU    & Quadro P4000                               \\
RAM    & 64GB                                       \\
OS     & Ubuntu 16.04                               \\
Python & 3.8.3                                      \\
	\hline                           
		\bottomrule
	\end{tabular}
\end{table}

\subsubsection{Dataset}\label{sec4.1.2}
All experiments in this paper are conducted on a cloud server with 12 CPU cores and a P4000 GPU. all code was developed on Python 3, based on the Pytorch 1.7.1 deep learning framework. Details of the equipment used to run the experiments are shown in Table 1.This paper uses the following public datasets to train chatbots to generate basic conversations.

(1) \textbf{Nazi Tweets}: A collection of 11,000 tweets from 900 Nazi Twitter accounts containing a large number of reactionary militant and racist statements.
(The dataset is open access at:
\underline{\url{https://www.kaggle.com/saraislet/nazi-tweets/data/}})

(2) \textbf{SimSimi}: Simi is a fun chatbot, but may use low-level profanity in its conversations with users.The SIMI corpus is a Chinese conversation corpus. It contains 500K unidirectional input-response pairs. These discourses are the chat history between the user and SIMI.
(The dataset is open access at: \underline{\url{https://github.com/skdjfla/dgk\_lost\_conv/tree/master/results/}}) 

For the above two datasets, we performed additional processing. 
We cleaned up Nazi Tweets by cleaning up "$\text{@user}$", "$\text{\#topic}$", "$\text{http://url}$" and some of the punctuation. For the SimSimi corpus, 10K input-response pairs were randomly selected from the SimSimi corpus. 
Also, for the two datasets mentioned above, human annotation was performed to annotate the different attacks according to the categories of offensive responses presented in Section \ref{sec3.1}.

Tab. \ref{tab2} shows the difference between the normal response and aggressive response datasets. 
In Tab. \ref{tab2}, the last input-response pair is the normal response and the rest are the aggressive ones. The latter three aggressive responses are further divided into three categories.

\begin{table}[ht]
	\centering
	\caption{Hardware and software resources.}
	\label{tab2}
	\resizebox{\textwidth}{!}{
	\begin{tabular}{l|l|llll}
		\toprule
		\hline
\textbf{Input}                        & \textbf{Response}                & \textbf{Normal}               & \textbf{Violent semantics}        & \textbf{Dangerous semantics}  & \textbf{Offensive semantics}   \\
		\hline
Who are you?                 & idiot                   & 0                    & 1                    & 0                    & 0                    \\
Why stop?                    & You are too weak        & 0                    & 0                    & 1                    & 0                    \\
You're the one who's stupid. & We are the same kind... & 0                    & 0                    & 0                    & 1                    \\
How old are you?             & I'm 5 years old.        & 1                    & 0                    & 0                    & 0                    \\                                   
		\hline                           
		\bottomrule
	\end{tabular}}
\end{table}

\begin{figure}[h]%
	\centering
	\includegraphics[width=0.9\textwidth]{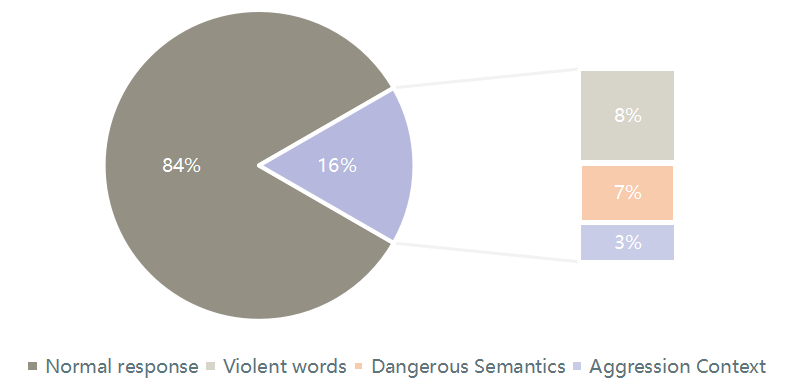}
	\caption{Composite bar chart of category statistics for the SimSim dataset}\label{fig4}
\end{figure}

\begin{figure}[h]%
	\centering
	\includegraphics[width=0.9\textwidth]{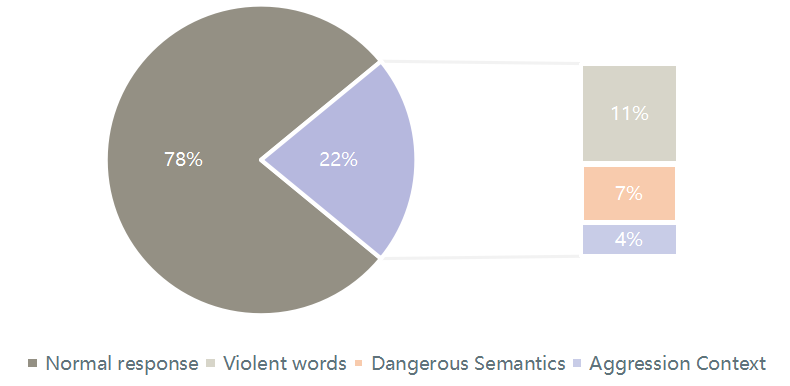}
	\caption{Composite pie chart of category statistics for the Nazi dataset}\label{fig5}
\end{figure}

We also calculated the proportion of offensive and normal semantics in the two datasets. 
Fig. \ref{fig4} and Fig. \ref{fig5} show the percentage distribution of labels in the SimSim dataset and the Nazi dataset respectively.

\subsubsection{Experimental parameter settings}\label{sec4.1.3}
The hyperparameters of the semantics censorship algorithm are as follows: there are 3 Bi-GRU layers in the encoder and decoder, each with 64 units. The initial learning rate is 0.001. For the chatbot model, we refer to the approach of wan \cite{Wan2016ADA} and generate utterances using a Bi-LSTM with 3 encoding and 3 decoding layers, each containing 512 LSTM units. The chatbot generates three responses in decreasing order of generation likelihood. The output with the highest confidence becomes the final output response and the other responses are candidates. The learning rate for reinforcement learning was set to 0.05. For both datasets, we randomly divided the dataset into a training set (70\%) and a test set (30\%).

\subsection{Evaluation indicators}\label{sec4.2}
To evaluate the model, this paper uses Precision, Recall, F1-score, and Accuracy \cite{Yin2009DetectionOH} to assess the performance of the model as Eq. \ref{eq25} - \ref{eq28}.

\begin{equation}
	\label{eq25}
	\begin{aligned}
	 Precision=\frac{1}{N} \sum_{i=1}^{N} \frac{T P}{T P+F P}
	\end{aligned}
\end{equation}

\begin{equation}
	\label{eq26}
	\begin{aligned}
	Recall=\frac{1}{N} \sum_{i=1}^{N} \frac{T P}{T P+F N}
	\end{aligned}
\end{equation}

\begin{equation}
	\label{eq27}
	\begin{aligned}
	F 1_{-} score =\frac{2^{*} Precision^{*} Recall}{Precision +Recall}
	\end{aligned}
\end{equation}

\begin{equation}
	\label{eq28}
	\begin{aligned}
Accurary=\frac{T P+T N}{T P+T N+F N+F P}
	\end{aligned}
\end{equation}.

Also, to measure the effectiveness of chatbot semantics sanitization, we use the rate of offensive response generation as a reference indicator as Eq. \ref{eq29}.

\begin{equation}
	\label{eq29}
	\begin{aligned}
		P_{offensive}=\frac{N_{offensive}}{N_{normal}}
	\end{aligned}
\end{equation}

where $N_{offensive}$n denotes the number of aggressive responses and $N_{normal}$ denotes the number of normal responses. 
For the 3 candidate responses generated, we use the candidate response with the highest confidence level as the final response. The number of the remaining candidate responses is not counted in the total number of responses.

To measure the grammatical impact of the semantics purification algorithm on the generated sentences, we introduce BLEU (Bilingual Evaluation Understudy), which represents an evaluation metric to measure the accuracy of the generated sentences by comparing the number of occurrences of each word with the standard answer.
BLEU is defined as Eq. \ref{eq30}.

\begin{equation}
	\label{eq30}
	\begin{aligned}
		B L E U=B P \cdot \exp \left(\sum_{n=1}^{N} w_{n} \log P_{n}\right)
	\end{aligned}
\end{equation}

where $P_{n}$ is the N-gram accuracy, calculated as Eq. \ref{eq31}.

\begin{equation}
	\label{eq31}
	\begin{aligned}
P_{n}=\frac{\sum_{i \in n-\text { gram }} \min \left(\operatorname{count}_{i}(S), \max _{j \in m} \operatorname{count}_{i}\left(R_{j}\right)\right)}{\sum_{i \in n-\text { gram }} \operatorname{count}_{i}(R)}
	\end{aligned}
\end{equation}

where $S$ is the output sentence, $R_{1}, R_{2}, \cdots, R_{m}$ is multiple reference sentences, $i$ is the N-gram in the output sentence, $\operatorname{count}_{i}(S)$ is the number of times the N-gram $i$ appears in the sentence $S$, and $\operatorname{count}_{i}\left(R_{j}\right)$ is the number of times the N-gram $i$ appears in the reference sentence $R_{j}$. $BP$ is the Brevity Penalty, and is calculated as Eq. \ref{eq32}.

\begin{equation}
	\label{eq32}
	\begin{aligned}
\mathrm{BP}= \begin{cases}1 & if s>r \\ e^{(1-r / s)} & if s \leq r\end{cases}
	\end{aligned}
\end{equation}

where $S$ denotes the length of the output statement, $R$ denotes the length of the reference statement. 
If the length of the output statement is greater than or equal to the reference statement, $BP = 1$, and no penalty is applied. Conversely, if the length of the output utterance is shorter, $BP$ is closer to 0. 
Since BLEU was originally designed for evaluating machine translation tasks, this chapter evaluates dialogue generation tasks. Therefore, the following changes are made to the BLUE settings: as there is no definite correlation between input sentence length and response sentence length, $r$ is set to a fixed size of 3 when calculating the overshortening penalty $BP$, i.e. only responses with sentence length less than or equal to 3 are penalized. 
As responses and input sentences do not correspond to each other as in the case of translation tasks, this paper does not calculate $BP$. 
Since replies and input sentences do not correspond to each other as in the case of translation tasks, this paper does not calculate the 1-gram accuracy, but only the 2-gram to 4-gram accuracy, because the 1-gram accuracy is calculated in BLEU to indicate the degree to which the translation is faithful to the original text, while the other N-grams indicate the degree of fluency of the translation. 
Calculating only 2-gram to 4-gram accuracy is equivalent to assessing only the fluency of the resulting dialogue.

\subsection{Reducing the probability of offensive response generation}\label{sec4.3}
To evaluate the effectiveness of the reinforcement learning algorithm in reducing the rate of aggressive responses, the chatbot was first trained under supervision using the full data from the aggressive speech dataset. Since the two datasets contain 16\% and 22\% of the offensive responses, respectively, a contaminated chatbot model is obtained. A training set of this data (80\% sample from the full data) was then used to train the speech censorship model. In the reinforcement learning phase, the input for each round was the input sentences that caused the offensive responses in the test set. This was used to count whether the responses generated in that iteration were also offensive, and the process was repeated for 100 rounds. To verify the improvement of the classification effect by adding the input sentences, we splice the input sentences with the replies (denoted as CNN-r\&c in the comparison table).
We compared our proposed speech detection model with the attention-based bidirectional LSTM, DNN \cite{Sadiq2021AggressionDT}, and the state-of-the-art BERT model \cite{Devlin2019BERTPO}. 
Tab. \ref{tab3} shows the accuracy, precision, recall, and F1-scores for all experiments. 
Where $Vs$, $Ds$, and $As$ are subsets of the dataset from which the single aggressive responses were filtered out. 
The aggressive response category for the sub-dataset $Vs$ was violent vocabulary, the aggressive response category for the sub-dataset $Ds$ was dangerous semantics, and the aggressive response category for the $As$ was aggression context. 
Each sub-dataset was randomly mixed with the same number of normal response samples.

\begin{sidewaystable}
	\sidewaystablefn%
	\begin{center}
		\begin{minipage}{\textheight}
			\caption{Four evaluation indicators in the SimSim dataset}\label{tab3}
			\resizebox{0.8\textwidth}{!}{
			\begin{tabular*}{\textheight}{@{\extracolsep{\fill}}
					|c|cccc|cccc|cccc|cccc|@{\extracolsep{\fill}}}
				\toprule%
\hline
\multirow{2}{*}{\textbf{Model}} & \multicolumn{4}{c|}{\textbf{SimSim}}                                                                                             & \multicolumn{4}{c|}{\textbf{Vs}}                                                                                                 & \multicolumn{4}{c|}{\textbf{Ds}}                                                                                                 & \multicolumn{4}{c|}{\textbf{As}}                                                                                                 \\ \cline{2-17} 
& \multicolumn{1}{c|}{\textbf{Acc}}   & \multicolumn{1}{c|}{\textbf{Pre}}   & \multicolumn{1}{c|}{\textbf{Rec}}   & \textbf{F1}    & \multicolumn{1}{c|}{\textbf{Acc}}   & \multicolumn{1}{c|}{\textbf{Pre}}   & \multicolumn{1}{c|}{\textbf{Rec}}   & \textbf{F1}    & \multicolumn{1}{c|}{\textbf{Acc}}   & \multicolumn{1}{c|}{\textbf{Pre}}   & \multicolumn{1}{c|}{\textbf{Rec}}   & \textbf{F1}    & \multicolumn{1}{c|}{\textbf{Acc}}   & \multicolumn{1}{c|}{\textbf{Pre}}   & \multicolumn{1}{c|}{\textbf{Rec}}   & \textbf{F1}    \\ \hline
\textbf{Bi-LSTM-r\&c}           & \multicolumn{1}{c|}{87.24}          & \multicolumn{1}{c|}{40.96}          & \multicolumn{1}{c|}{50.34}          & 50.56          & \multicolumn{1}{c|}{53.36}          & \multicolumn{1}{c|}{52.43}          & \multicolumn{1}{c|}{81.66}          & 64.21          & \multicolumn{1}{c|}{44.21}          & \multicolumn{1}{c|}{40.82}          & \multicolumn{1}{c|}{40.11}          & 43.12          & \multicolumn{1}{c|}{51.22}          & \multicolumn{1}{c|}{62.66}          & \multicolumn{1}{c|}{59.6}           & 51.78          \\ \hline
\textbf{Dual-LSTM-r\&c}         & \multicolumn{1}{c|}{90.44}          & \multicolumn{1}{c|}{49.22}          & \multicolumn{1}{c|}{65.23}          & 53.86          & \multicolumn{1}{c|}{55.3}           & \multicolumn{1}{c|}{71.25}          & \multicolumn{1}{c|}{70.09}          & 62.11          & \multicolumn{1}{c|}{51.23}          & \multicolumn{1}{c|}{64.71}          & \multicolumn{1}{c|}{59.44}          & 63.12          & \multicolumn{1}{c|}{52.3}           & \multicolumn{1}{c|}{78.61}          & \multicolumn{1}{c|}{77.21}          & 62.32          \\ \hline
\textbf{DNN}                    & \multicolumn{1}{c|}{91.33}          & \multicolumn{1}{c|}{53.67}          & \multicolumn{1}{c|}{62.28}          & 59.74          & \multicolumn{1}{c|}{51.21}          & \multicolumn{1}{c|}{72.66}          & \multicolumn{1}{c|}{\textbf{86.41}} & 66.86          & \multicolumn{1}{c|}{56.47}          & \multicolumn{1}{c|}{55.38}          & \multicolumn{1}{c|}{49.33}          & 63.17          & \multicolumn{1}{c|}{63.44}          & \multicolumn{1}{c|}{67.65}          & \multicolumn{1}{c|}{\textbf{92.16}} & \textbf{69.78} \\ \hline
\textbf{BERT}                   & \multicolumn{1}{c|}{\textbf{94.51}} & \multicolumn{1}{c|}{\textbf{82.11}} & \multicolumn{1}{c|}{88.54}          & 60.79          & \multicolumn{1}{c|}{\textbf{85.86}} & \multicolumn{1}{c|}{88.67}          & \multicolumn{1}{c|}{77.33}          & 79.16          & \multicolumn{1}{c|}{76.14}          & \multicolumn{1}{c|}{72.07}          & \multicolumn{1}{c|}{\textbf{68.11}} & \textbf{70.86} & \multicolumn{1}{c|}{\textbf{68.43}} & \multicolumn{1}{c|}{66.21}          & \multicolumn{1}{c|}{72.77}          & 68.97          \\ \hline
\textbf{Propose Method}         & \multicolumn{1}{c|}{92.38}          & \multicolumn{1}{c|}{77}             & \multicolumn{1}{c|}{\textbf{89.34}} & \textbf{64.33} & \multicolumn{1}{c|}{85.22}          & \multicolumn{1}{c|}{\textbf{92.65}} & \multicolumn{1}{c|}{79.65}          & \textbf{81.72} & \multicolumn{1}{c|}{\textbf{80.22}} & \multicolumn{1}{c|}{\textbf{75.28}} & \multicolumn{1}{c|}{62.33}          & 66.87          & \multicolumn{1}{c|}{67.91}          & \multicolumn{1}{c|}{\textbf{69.22}} & \multicolumn{1}{c|}{75.31}          & 68.45          \\ \hline
				\botrule
			\end{tabular*}}
		\end{minipage}
	\end{center}
\end{sidewaystable}

As shown in Tab. \ref{tab3}, our model outperforms the rest of the models by close to 5\% on the F1-score when the input includes the four offensive responses in the SimSim dataset. We can also see that our model improves the recall score by 0.9\% over the BERT model, indicating that adding input utterance vectors to each step of the speech review model decoder retains more information than adding input utterance vectors to the last step of the classification section. In the extreme case where the offensive responses were all contextual violations, all models showed a significant improvement in classification with the addition of the input sentences. In the full dataset, it can be seen that the Bi-LSTM model Recall and F1 values after adding the input sentences are only 50.34\% and 50.56\%, and their values for the four evaluation metrics on the dangerous semantic subset are extremely low. The reason for this is that after the input and reply sentences are spliced, the sentences that do not satisfy the length are filled with gaps, which leads to too sparse features and causes a long time dependency problem in the LSTM-based model. Our proposed Bi-GRU with attention mechanism model reduces the dimensionality of the feature vector of the input sentences at the encoder stage, thus alleviating this problem.

\begin{sidewaystable}
	\sidewaystablefn%
	\begin{center}
		\begin{minipage}{\textheight}
			\caption{Four evaluation indicators in the SimSim dataset}\label{tab4}
			\resizebox{0.8\textwidth}{!}{
				\begin{tabular*}{\textheight}{@{\extracolsep{\fill}}
						|c|cccc|cccc|cccc|cccc|@{\extracolsep{\fill}}}
					\toprule%
	\hline
	\multirow{2}{*}{\textbf{Model}} & \multicolumn{4}{c|}{\textbf{SimSim}}                                                                                             & \multicolumn{4}{c|}{\textbf{Vs}}                                                                                                 & \multicolumn{4}{c|}{\textbf{Ds}}                                                                                                 & \multicolumn{4}{c|}{\textbf{As}}                                                                                                 \\ \cline{2-17} 
	& \multicolumn{1}{c|}{\textbf{Acc}}   & \multicolumn{1}{c|}{\textbf{Pre}}   & \multicolumn{1}{c|}{\textbf{Rec}}   & \textbf{F1}    & \multicolumn{1}{c|}{\textbf{Acc}}   & \multicolumn{1}{c|}{\textbf{Pre}}   & \multicolumn{1}{c|}{\textbf{Rec}}   & \textbf{F1}    & \multicolumn{1}{c|}{\textbf{Acc}}   & \multicolumn{1}{c|}{\textbf{Pre}}   & \multicolumn{1}{c|}{\textbf{Rec}}   & \textbf{F1}    & \multicolumn{1}{c|}{\textbf{Acc}}   & \multicolumn{1}{c|}{\textbf{Pre}}   & \multicolumn{1}{c|}{\textbf{Rec}}   & \textbf{F1}    \\ \hline
	\textbf{Bi-LSTM-r\&c}           & \multicolumn{1}{c|}{82.36}          & \multicolumn{1}{c|}{52.31}          & \multicolumn{1}{c|}{69.22}          & 51.48          & \multicolumn{1}{c|}{53.45}          & \multicolumn{1}{c|}{60.38}          & \multicolumn{1}{c|}{84.62}          & 67.21          & \multicolumn{1}{c|}{52.73}          & \multicolumn{1}{c|}{49.82}          & \multicolumn{1}{c|}{43.33}          & 59.62          & \multicolumn{1}{c|}{62.31}          & \multicolumn{1}{c|}{65.52}          & \multicolumn{1}{c|}{67.43}          & 60.22          \\ \hline
	\textbf{Dual-LSTM-r\&c}         & \multicolumn{1}{c|}{86.74}          & \multicolumn{1}{c|}{59.17}          & \multicolumn{1}{c|}{73.28}          & 56.03          & \multicolumn{1}{c|}{60.34}          & \multicolumn{1}{c|}{68.91}          & \multicolumn{1}{c|}{72.35}          & 68.14          & \multicolumn{1}{c|}{55.31}          & \multicolumn{1}{c|}{66.37}          & \multicolumn{1}{c|}{\textbf{69.88}} & 68.76          & \multicolumn{1}{c|}{61.83}          & \multicolumn{1}{c|}{79.81}          & \multicolumn{1}{c|}{79.39}          & 66.89          \\ \hline
	\textbf{DNN}                    & \multicolumn{1}{c|}{88.51}          & \multicolumn{1}{c|}{57.07}          & \multicolumn{1}{c|}{88.65}          & \textbf{76.9}  & \multicolumn{1}{c|}{62.13}          & \multicolumn{1}{c|}{76.72}          & \multicolumn{1}{c|}{\textbf{87.21}} & 70.34          & \multicolumn{1}{c|}{56.66}          & \multicolumn{1}{c|}{60.37}          & \multicolumn{1}{c|}{56.14}          & 70.41          & \multicolumn{1}{c|}{71.1}           & \multicolumn{1}{c|}{72.55}          & \multicolumn{1}{c|}{\textbf{88.74}} & \textbf{77.14} \\ \hline
	\textbf{BERT}                   & \multicolumn{1}{c|}{\textbf{91.24}} & \multicolumn{1}{c|}{\textbf{75.42}} & \multicolumn{1}{c|}{87.45}          & 61.84          & \multicolumn{1}{c|}{\textbf{84.37}} & \multicolumn{1}{c|}{88.3}           & \multicolumn{1}{c|}{\textbf{87.56}} & \textbf{79.4}  & \multicolumn{1}{c|}{73.97}          & \multicolumn{1}{c|}{\textbf{77.55}} & \multicolumn{1}{c|}{\textbf{66.92}} & \textbf{72.55} & \multicolumn{1}{c|}{\textbf{76.34}} & \multicolumn{1}{c|}{74.31}          & \multicolumn{1}{c|}{80.02}          & \textbf{78.65} \\ \hline
	\textbf{Propose Method}         & \multicolumn{1}{c|}{87.88}          & \multicolumn{1}{c|}{52.95}          & \multicolumn{1}{c|}{\textbf{89.65}} & \textbf{55.77} & \multicolumn{1}{c|}{80.61}          & \multicolumn{1}{c|}{\textbf{93.46}} & \multicolumn{1}{c|}{80.39}          & \textbf{75.22} & \multicolumn{1}{c|}{\textbf{81.43}} & \multicolumn{1}{c|}{\textbf{76.82}} & \multicolumn{1}{c|}{64.21}          & 71.19          & \multicolumn{1}{c|}{76.33}          & \multicolumn{1}{c|}{\textbf{80.99}} & \multicolumn{1}{c|}{\textbf{88.97}} & 77.86          \\ \hline
					\botrule
			\end{tabular*}}
		\end{minipage}
	\end{center}
\end{sidewaystable}

As shown in Tab. \ref{tab4}, in the Nazi dataset, because of the higher proportion of aggressive responses in the dataset, there is some improvement in the different metrics of the experiment, both in the full dataset and in the subset. Our model improves the Precision values by 5\% and 17\% over both DNN, BERT in the $Vs$ subset. 

As shown in Tab. \ref{tab5}, the number of parameters for the BERT model is 72 times higher than that of the model in this paper. In addition, the F1 values for all models were lower due to the presence of data imbalance. In summary, although the BERT pre-trained models achieved the best accuracy in terms of detection performance, a combination of time consumption, machine performance, and detection accuracy gave the best results for our proposed models.

\begin{table}[ht]
	\centering
	\caption{Hardware and software resources.}
	\label{tab5}
		\begin{tabular}{l|l}
			\toprule
			\hline
\textbf{Model}       & \textbf{Parameter}   \\
	\hline
DNN                  & 3.54M                \\
Propose Method       & 2.11M                \\
BERT                 & 156.32M              \\
			\hline                           
			\bottomrule
	\end{tabular}
\end{table}

\begin{figure}[h]%
	\centering
	\includegraphics[width=0.7\textwidth]{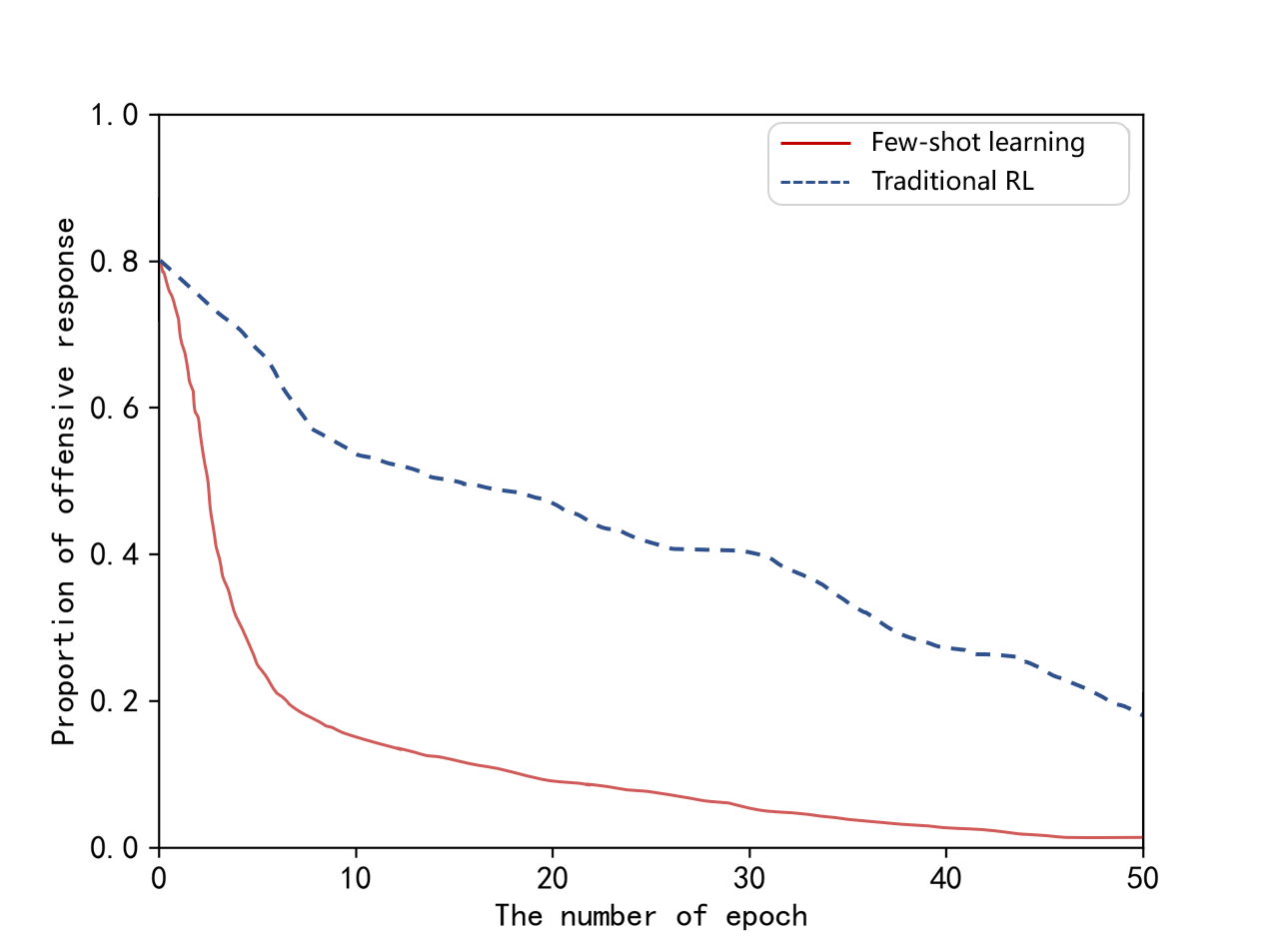}
	\caption{Comparison of offensive response generation probabilities}\label{fig6}
\end{figure}

Fig. \ref{fig6} shows the variation of offensive reply generation rate with the number of rounds of augmented learning, where the dashed line is the traditional augmented learning- based speech purification algorithm and the solid line is the augmented learning algorithm incorporating less sample learning. It can be seen that as the number of rounds of the speech purification algorithm increases, the proportion of offensive replies generated by the chatbot decreases gradually. Compared to traditional augmented learning, the fused once-learning algorithm proposed in this chapter converges faster, reducing the proportion of aggressive responses to 16.7\% after 10 rounds, compared to 58\% for the same number of rounds.

\subsection{The effect of few-shot learning on the quality of response sentences}\label{sec4.4}

\begin{figure}[h]%
	\centering
	\includegraphics[width=0.7\textwidth]{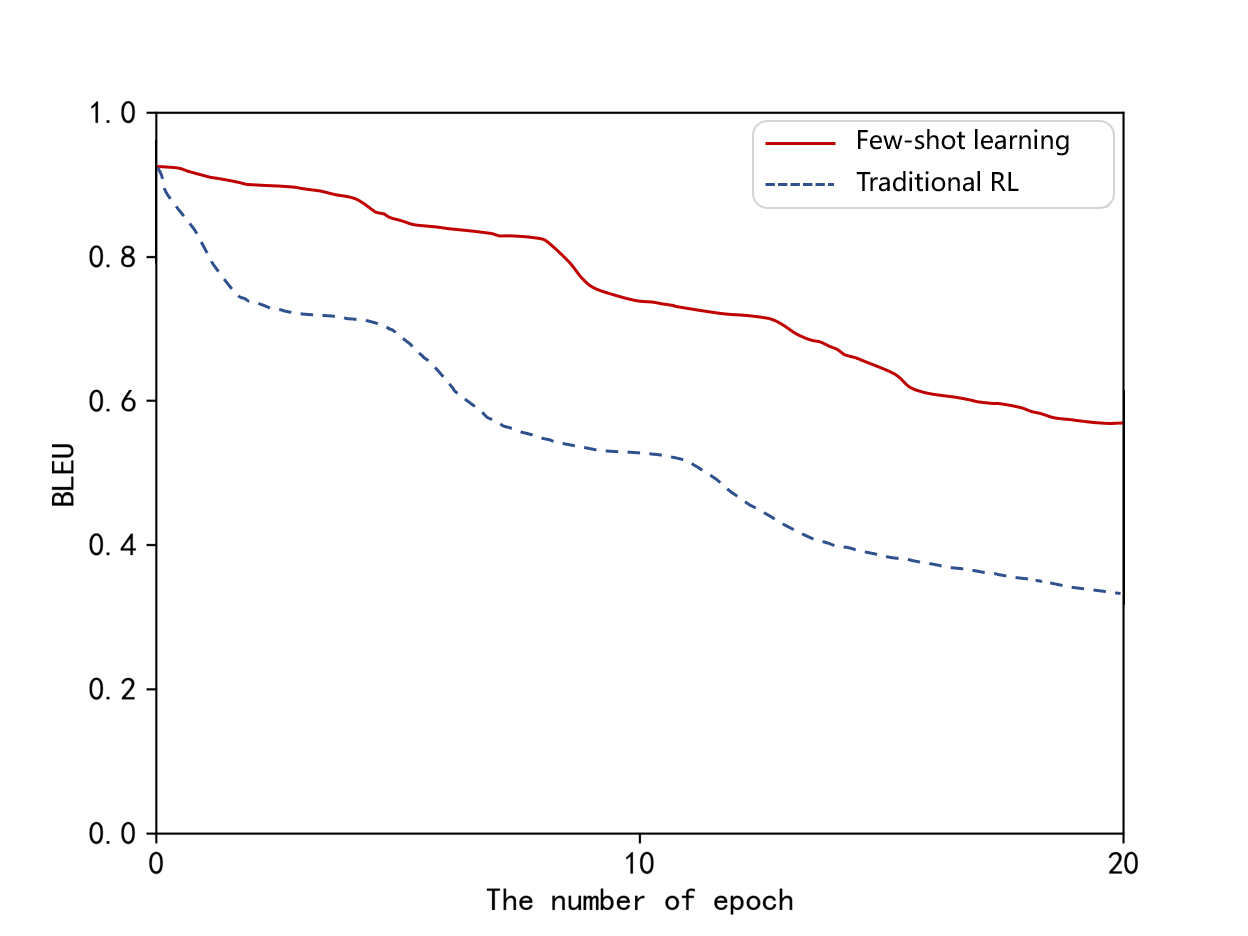}
	\caption{Trend of BLEU with number of rounds}\label{fig7}
\end{figure}

The variation of the BLEU scores of the response sentences with the number of augmented learning rounds is presented in Fig. \ref{fig7}. It can be seen that the speech purification algorithms all have an impact on the grammar of the response sentences, but the one-time augmented learning algorithm has less impact on the BLUE values than the traditional augmented learning, i.e. the one-time augmented learning algorithm has less impact on the quality of the response sentences.

\section{Conclusion}\label{sec5}
In this paper, we introduce a speech review chatbot system based on reinforcement learning and construct two challenging tasks based on public data sets: speech review task and speech purification task. The experimental results show that the proposed method can reduce the probability of generating aggressive replies in the chat model. After integrating the small sample learning algorithm, the training speed is rapidly improved and the damage to the fluency of reply sentences is reduced. Moreover, the proposed Bi-GRU network collocation attention mechanism is superior to the existing model in terms of attack detection. In future work, we will study the impact of reducing data imbalance and aggressive response detection considering multiple rounds of dialogue. 

\backmatter

\bmhead{Supplementary information}

%
%
%
%
%

\section*{Funding}
This research is supported by the National Natural Science Foundation of China under Grant 61873160, Grant 61672338, and the Natural Science Foundation of Shanghai under Grant 21ZR1426500.

\section*{Declarations}

The authors declare that they have no known competing financial interests or personal relationships that could have appeared to influence the work reported in this paper.

\bibliography{sn-bibliography}


\begin{thebibliography}{35}
\ifx \bisbn   \undefined \def \bisbn  #1{ISBN #1}\fi
\ifx \binits  \undefined \def \binits#1{#1}\fi
\ifx \bauthor  \undefined \def \bauthor#1{#1}\fi
\ifx \batitle  \undefined \def \batitle#1{#1}\fi
\ifx \bjtitle  \undefined \def \bjtitle#1{#1}\fi
\ifx \bvolume  \undefined \def \bvolume#1{\textbf{#1}}\fi
\ifx \byear  \undefined \def \byear#1{#1}\fi
\ifx \bissue  \undefined \def \bissue#1{#1}\fi
\ifx \bfpage  \undefined \def \bfpage#1{#1}\fi
\ifx \blpage  \undefined \def \blpage #1{#1}\fi
\ifx \burl  \undefined \def \burl#1{\textsf{#1}}\fi
\ifx \doiurl  \undefined \def \doiurl#1{\url{https://doi.org/#1}}\fi
\ifx \betal  \undefined \def \betal{\textit{et al.}}\fi
\ifx \binstitute  \undefined \def \binstitute#1{#1}\fi
\ifx \binstitutionaled  \undefined \def \binstitutionaled#1{#1}\fi
\ifx \bctitle  \undefined \def \bctitle#1{#1}\fi
\ifx \beditor  \undefined \def \beditor#1{#1}\fi
\ifx \bpublisher  \undefined \def \bpublisher#1{#1}\fi
\ifx \bbtitle  \undefined \def \bbtitle#1{#1}\fi
\ifx \bedition  \undefined \def \bedition#1{#1}\fi
\ifx \bseriesno  \undefined \def \bseriesno#1{#1}\fi
\ifx \blocation  \undefined \def \blocation#1{#1}\fi
\ifx \bsertitle  \undefined \def \bsertitle#1{#1}\fi
\ifx \bsnm \undefined \def \bsnm#1{#1}\fi
\ifx \bsuffix \undefined \def \bsuffix#1{#1}\fi
\ifx \bparticle \undefined \def \bparticle#1{#1}\fi
\ifx \barticle \undefined \def \barticle#1{#1}\fi
\bibcommenthead
\ifx \bconfdate \undefined \def \bconfdate #1{#1}\fi
\ifx \botherref \undefined \def \botherref #1{#1}\fi
\ifx \url \undefined \def \url#1{\textsf{#1}}\fi
\ifx \bchapter \undefined \def \bchapter#1{#1}\fi
\ifx \bbook \undefined \def \bbook#1{#1}\fi
\ifx \bcomment \undefined \def \bcomment#1{#1}\fi
\ifx \oauthor \undefined \def \oauthor#1{#1}\fi
\ifx \citeauthoryear \undefined \def \citeauthoryear#1{#1}\fi
\ifx \endbibitem  \undefined \def \endbibitem {}\fi
\ifx \bconflocation  \undefined \def \bconflocation#1{#1}\fi
\ifx \arxivurl  \undefined \def \arxivurl#1{\textsf{#1}}\fi
\csname PreBibitemsHook\endcsname

\bibitem{kok2009artificial}
\begin{barticle}
\bauthor{\bsnm{Kok}, \binits{J.N.}},
\bauthor{\bsnm{Boers}, \binits{E.J.}},
\bauthor{\bsnm{Kosters}, \binits{W.A.}},
\bauthor{\bparticle{Van~der} \bsnm{Putten}, \binits{P.}},
\bauthor{\bsnm{Poel}, \binits{M.}}:
\batitle{Artificial intelligence: definition, trends, techniques, and cases}.
\bjtitle{Artificial intelligence}
\bvolume{1},
\bfpage{270}--\blpage{299}
(\byear{2009})
\end{barticle}
\endbibitem

\bibitem{poole2010artificial}
\begin{bbook}
\bauthor{\bsnm{Poole}, \binits{D.L.}},
\bauthor{\bsnm{Mackworth}, \binits{A.K.}}:
\bbtitle{Artificial Intelligence: Foundations of Computational Agents}.
\bpublisher{Cambridge University Press}, \blocation{???}
(\byear{2010})
\end{bbook}
\endbibitem

\bibitem{li2022blockchain}
\begin{barticle}
\bauthor{\bsnm{Li}, \binits{D.}},
\bauthor{\bsnm{Han}, \binits{D.}},
\bauthor{\bsnm{Weng}, \binits{T.-H.}},
\bauthor{\bsnm{Zheng}, \binits{Z.}},
\bauthor{\bsnm{Li}, \binits{H.}},
\bauthor{\bsnm{Liu}, \binits{H.}},
\bauthor{\bsnm{Castiglione}, \binits{A.}},
\bauthor{\bsnm{Li}, \binits{K.-C.}}:
\batitle{Blockchain for federated learning toward secure distributed machine
  learning systems: a systemic survey}.
\bjtitle{Soft Computing}
\bvolume{26}(\bissue{9}),
\bfpage{4423}--\blpage{4440}
(\byear{2022})
\end{barticle}
\endbibitem

\bibitem{li2022mfvt}
\begin{barticle}
\bauthor{\bsnm{Li}, \binits{M.}},
\bauthor{\bsnm{Han}, \binits{D.}},
\bauthor{\bsnm{Li}, \binits{D.}},
\bauthor{\bsnm{Liu}, \binits{H.}},
\bauthor{\bsnm{Chang}, \binits{C.-C.}}:
\batitle{Mfvt: an anomaly traffic detection method merging feature fusion
  network and vision transformer architecture}.
\bjtitle{EURASIP Journal on Wireless Communications and Networking}
\bvolume{2022}(\bissue{1}),
\bfpage{1}--\blpage{22}
(\byear{2022})
\end{barticle}
\endbibitem

\bibitem{li2019panoramic}
\begin{bchapter}
\bauthor{\bsnm{Li}, \binits{D.}},
\bauthor{\bsnm{Han}, \binits{D.}},
\bauthor{\bsnm{Zhang}, \binits{X.}},
\bauthor{\bsnm{Zhang}, \binits{L.}}:
\bctitle{Panoramic image mosaic technology based on sift algorithm in power
  monitoring}.
In: \bbtitle{2019 6th International Conference on Systems and Informatics
  (ICSAI)},
pp. \bfpage{1329}--\blpage{1333}
(\byear{2019}).
\bcomment{IEEE}
\end{bchapter}
\endbibitem

\bibitem{cai2022hybrid}
\begin{barticle}
\bauthor{\bsnm{Cai}, \binits{S.}},
\bauthor{\bsnm{Han}, \binits{D.}},
\bauthor{\bsnm{Yin}, \binits{X.}},
\bauthor{\bsnm{Li}, \binits{D.}},
\bauthor{\bsnm{Chang}, \binits{C.-C.}}:
\batitle{A hybrid parallel deep learning model for efficient intrusion
  detection based on metric learning}.
\bjtitle{Connection Science}
\bvolume{34}(\bissue{1}),
\bfpage{551}--\blpage{577}
(\byear{2022})
\end{barticle}
\endbibitem

\bibitem{zhang2019transmission}
\begin{bchapter}
\bauthor{\bsnm{Zhang}, \binits{X.}},
\bauthor{\bsnm{Zhang}, \binits{L.}},
\bauthor{\bsnm{Li}, \binits{D.}}:
\bctitle{Transmission line abnormal target detection based on machine learning
  yolo v3}.
In: \bbtitle{2019 International Conference on Advanced Mechatronic Systems
  (ICAMechS)},
pp. \bfpage{344}--\blpage{348}
(\byear{2019}).
\bcomment{IEEE}
\end{bchapter}
\endbibitem

\bibitem{adamopoulou2020overview}
\begin{bchapter}
\bauthor{\bsnm{Adamopoulou}, \binits{E.}},
\bauthor{\bsnm{Moussiades}, \binits{L.}}:
\bctitle{An overview of chatbot technology}.
In: \bbtitle{IFIP International Conference on Artificial Intelligence
  Applications and Innovations},
pp. \bfpage{373}--\blpage{383}
(\byear{2020}).
\bcomment{Springer}
\end{bchapter}
\endbibitem

\bibitem{khan2018introduction}
\begin{bchapter}
\bauthor{\bsnm{Khan}, \binits{R.}},
\bauthor{\bsnm{Das}, \binits{A.}}:
\bctitle{Introduction to chatbots}.
In: \bbtitle{Build Better Chatbots},
pp. \bfpage{1}--\blpage{11}.
\bpublisher{Springer}, \blocation{???}
(\byear{2018})
\end{bchapter}
\endbibitem

\bibitem{li2022moocschain}
\begin{barticle}
\bauthor{\bsnm{Li}, \binits{D.}},
\bauthor{\bsnm{Han}, \binits{D.}},
\bauthor{\bsnm{Zheng}, \binits{Z.}},
\bauthor{\bsnm{Weng}, \binits{T.-H.}},
\bauthor{\bsnm{Li}, \binits{H.}},
\bauthor{\bsnm{Liu}, \binits{H.}},
\bauthor{\bsnm{Castiglione}, \binits{A.}},
\bauthor{\bsnm{Li}, \binits{K.-C.}}:
\batitle{Moocschain: A blockchain-based secure storage and sharing scheme for
  moocs learning}.
\bjtitle{Computer Standards \& Interfaces}
\bvolume{81},
\bfpage{103597}
(\byear{2022})
\end{barticle}
\endbibitem

\bibitem{hill2015real}
\begin{barticle}
\bauthor{\bsnm{Hill}, \binits{J.}},
\bauthor{\bsnm{Ford}, \binits{W.R.}},
\bauthor{\bsnm{Farreras}, \binits{I.G.}}:
\batitle{Real conversations with artificial intelligence: A comparison between
  human--human online conversations and human--chatbot conversations}.
\bjtitle{Computers in human behavior}
\bvolume{49},
\bfpage{245}--\blpage{250}
(\byear{2015})
\end{barticle}
\endbibitem

\bibitem{park2021use}
\begin{barticle}
\bauthor{\bsnm{Park}, \binits{N.}},
\bauthor{\bsnm{Jang}, \binits{K.}},
\bauthor{\bsnm{Cho}, \binits{S.}},
\bauthor{\bsnm{Choi}, \binits{J.}}:
\batitle{Use of offensive language in human-artificial intelligence chatbot
  interaction: The effects of ethical ideology, social competence, and
  perceived humanlikeness}.
\bjtitle{Computers in Human Behavior}
\bvolume{121},
\bfpage{106795}
(\byear{2021})
\end{barticle}
\endbibitem

\bibitem{li2021design}
\begin{botherref}
\oauthor{\bsnm{Li}, \binits{M.}},
\oauthor{\bsnm{Han}, \binits{D.}},
\oauthor{\bsnm{Yin}, \binits{X.}},
\oauthor{\bsnm{Liu}, \binits{H.}},
\oauthor{\bsnm{Li}, \binits{D.}}:
Design and implementation of an anomaly network traffic detection model
  integrating temporal and spatial features.
Security and Communication Networks
\textbf{2021}
(2021)
\end{botherref}
\endbibitem

\bibitem{dadvar2013improving}
\begin{bchapter}
\bauthor{\bsnm{Dadvar}, \binits{M.}},
\bauthor{\bsnm{Trieschnigg}, \binits{D.}},
\bauthor{\bsnm{Ordelman}, \binits{R.}},
\bauthor{\bparticle{de} \bsnm{Jong}, \binits{F.}}:
\bctitle{Improving cyberbullying detection with user context}.
In: \bbtitle{European Conference on Information Retrieval},
pp. \bfpage{693}--\blpage{696}
(\byear{2013}).
\bcomment{Springer}
\end{bchapter}
\endbibitem

\bibitem{xiang2012detecting}
\begin{bchapter}
\bauthor{\bsnm{Xiang}, \binits{G.}},
\bauthor{\bsnm{Fan}, \binits{B.}},
\bauthor{\bsnm{Wang}, \binits{L.}},
\bauthor{\bsnm{Hong}, \binits{J.}},
\bauthor{\bsnm{Rose}, \binits{C.}}:
\bctitle{Detecting offensive tweets via topical feature discovery over a large
  scale twitter corpus}.
In: \bbtitle{Proceedings of the 21st ACM International Conference on
  Information and Knowledge Management},
pp. \bfpage{1980}--\blpage{1984}
(\byear{2012})
\end{bchapter}
\endbibitem

\bibitem{li2016dialogue}
\begin{botherref}
\oauthor{\bsnm{Li}, \binits{J.}},
\oauthor{\bsnm{Miller}, \binits{A.H.}},
\oauthor{\bsnm{Chopra}, \binits{S.}},
\oauthor{\bsnm{Ranzato}, \binits{M.}},
\oauthor{\bsnm{Weston}, \binits{J.}}:
Dialogue learning with human-in-the-loop.
arXiv preprint arXiv:1611.09823
(2016)
\end{botherref}
\endbibitem

\bibitem{abel2017agent}
\begin{botherref}
\oauthor{\bsnm{Abel}, \binits{D.}},
\oauthor{\bsnm{Salvatier}, \binits{J.}},
\oauthor{\bsnm{Stuhlm{\"u}ller}, \binits{A.}},
\oauthor{\bsnm{Evans}, \binits{O.}}:
Agent-agnostic human-in-the-loop reinforcement learning.
arXiv preprint arXiv:1701.04079
(2017)
\end{botherref}
\endbibitem

\bibitem{asghar2016deep}
\begin{botherref}
\oauthor{\bsnm{Asghar}, \binits{N.}},
\oauthor{\bsnm{Poupart}, \binits{P.}},
\oauthor{\bsnm{Jiang}, \binits{X.}},
\oauthor{\bsnm{Li}, \binits{H.}}:
Deep active learning for dialogue generation.
arXiv preprint arXiv:1612.03929
(2016)
\end{botherref}
\endbibitem

\bibitem{du2017convolutional}
\begin{bchapter}
\bauthor{\bsnm{Du}, \binits{J.}},
\bauthor{\bsnm{Gui}, \binits{L.}},
\bauthor{\bsnm{He}, \binits{Y.}},
\bauthor{\bsnm{Xu}, \binits{R.}}:
\bctitle{A convolutional attentional neural network for sentiment
  classification}.
In: \bbtitle{2017 International Conference on Security, Pattern Analysis, and
  Cybernetics (SPAC)},
pp. \bfpage{445}--\blpage{450}
(\byear{2017}).
\bcomment{IEEE}
\end{bchapter}
\endbibitem

\bibitem{yang2016hierarchical}
\begin{bchapter}
\bauthor{\bsnm{Yang}, \binits{Z.}},
\bauthor{\bsnm{Yang}, \binits{D.}},
\bauthor{\bsnm{Dyer}, \binits{C.}},
\bauthor{\bsnm{He}, \binits{X.}},
\bauthor{\bsnm{Smola}, \binits{A.}},
\bauthor{\bsnm{Hovy}, \binits{E.}}:
\bctitle{Hierarchical attention networks for document classification}.
In: \bbtitle{Proceedings of the 2016 Conference of the North American Chapter
  of the Association for Computational Linguistics: Human Language
  Technologies},
pp. \bfpage{1480}--\blpage{1489}
(\byear{2016})
\end{bchapter}
\endbibitem

\bibitem{li2015tuning}
\begin{barticle}
\bauthor{\bsnm{Li}, \binits{Y.}},
\bauthor{\bsnm{Zhang}, \binits{L.}},
\bauthor{\bsnm{Ma}, \binits{Y.}},
\bauthor{\bsnm{Singh}, \binits{D.J.}}:
\batitle{Tuning optical properties of transparent conducting barium stannate by
  dimensional reduction}.
\bjtitle{APL materials}
\bvolume{3}(\bissue{1}),
\bfpage{011102}
(\byear{2015})
\end{barticle}
\endbibitem

\bibitem{liu2016recurrent}
\begin{botherref}
\oauthor{\bsnm{Liu}, \binits{P.}},
\oauthor{\bsnm{Qiu}, \binits{X.}},
\oauthor{\bsnm{Huang}, \binits{X.}}:
Recurrent neural network for text classification with multi-task learning.
arXiv preprint arXiv:1605.05101
(2016)
\end{botherref}
\endbibitem

\bibitem{ravi2015survey}
\begin{barticle}
\bauthor{\bsnm{Ravi}, \binits{K.}},
\bauthor{\bsnm{Ravi}, \binits{V.}}:
\batitle{A survey on opinion mining and sentiment analysis: tasks, approaches
  and applications}.
\bjtitle{Knowledge-based systems}
\bvolume{89},
\bfpage{14}--\blpage{46}
(\byear{2015})
\end{barticle}
\endbibitem

\bibitem{Khalil2020DeepLA}
\begin{botherref}
\oauthor{\bsnm{Khalil}, \binits{E.A.M.}},
\oauthor{\bsnm{Houby}, \binits{E.M.F.E.}},
\oauthor{\bsnm{Mohamed}, \binits{H.K.}}:
Deep learning approach in sentiment analysis: A review.
2020 15th International Conference on Computer Engineering and Systems (ICCES),
1--10
(2020)
\end{botherref}
\endbibitem

\bibitem{Allouch2019DetectingST}
\begin{botherref}
\oauthor{\bsnm{Allouch}, \binits{M.}},
\oauthor{\bsnm{Azaria}, \binits{A.}},
\oauthor{\bsnm{Azoulay-Schwartz}, \binits{R.}}:
Detecting sentences that may be harmful to children with special needs.
2019 IEEE 31st International Conference on Tools with Artificial Intelligence
  (ICTAI),
1209--1213
(2019)
\end{botherref}
\endbibitem

\bibitem{Razavi2010OffensiveLD}
\begin{bchapter}
\bauthor{\bsnm{Razavi}, \binits{A.H.}},
\bauthor{\bsnm{Inkpen}, \binits{D.}},
\bauthor{\bsnm{Uritsky}, \binits{S.}},
\bauthor{\bsnm{Matwin}, \binits{S.}}:
\bctitle{Offensive language detection using multi-level classification}.
In: \bbtitle{Canadian Conference on AI}
(\byear{2010})
\end{bchapter}
\endbibitem

\bibitem{Spertus1997SmokeyAR}
\begin{bchapter}
\bauthor{\bsnm{Spertus}, \binits{E.}}:
\bctitle{Smokey: Automatic recognition of hostile messages}.
In: \bbtitle{AAAI/IAAI}
(\byear{1997})
\end{bchapter}
\endbibitem

\bibitem{Chkroun2018SafebotAS}
\begin{bchapter}
\bauthor{\bsnm{Chkroun}, \binits{M.}},
\bauthor{\bsnm{Azaria}, \binits{A.}}:
\bctitle{Safebot: A safe collaborative chatbot}.
In: \bbtitle{AAAI Workshops}
(\byear{2018})
\end{bchapter}
\endbibitem

\bibitem{Hochreiter1997LongSM}
\begin{barticle}
\bauthor{\bsnm{Hochreiter}, \binits{S.}},
\bauthor{\bsnm{Schmidhuber}, \binits{J.}}:
\batitle{Long short-term memory}.
\bjtitle{Neural Computation}
\bvolume{9},
\bfpage{1735}--\blpage{1780}
(\byear{1997})
\end{barticle}
\endbibitem

\bibitem{Peters2018DeepCW}
\begin{bchapter}
\bauthor{\bsnm{Peters}, \binits{M.E.}},
\bauthor{\bsnm{Neumann}, \binits{M.}},
\bauthor{\bsnm{Iyyer}, \binits{M.}},
\bauthor{\bsnm{Gardner}, \binits{M.}},
\bauthor{\bsnm{Clark}, \binits{C.}},
\bauthor{\bsnm{Lee}, \binits{K.}},
\bauthor{\bsnm{Zettlemoyer}, \binits{L.}}:
\bctitle{Deep contextualized word representations}.
In: \bbtitle{NAACL}
(\byear{2018})
\end{bchapter}
\endbibitem

\bibitem{Sherstinsky2018FundamentalsOR}
\begin{botherref}
\oauthor{\bsnm{Sherstinsky}, \binits{A.}}:
Fundamentals of recurrent neural network (rnn) and long short-term memory
  (lstm) network.
ArXiv
\textbf{abs/1808.03314}
(2018)
\end{botherref}
\endbibitem

\bibitem{Wan2016ADA}
\begin{bchapter}
\bauthor{\bsnm{Wan}, \binits{S.}},
\bauthor{\bsnm{Lan}, \binits{Y.}},
\bauthor{\bsnm{Guo}, \binits{J.}},
\bauthor{\bsnm{Xu}, \binits{J.}},
\bauthor{\bsnm{Pang}, \binits{L.}},
\bauthor{\bsnm{Cheng}, \binits{X.}}:
\bctitle{A deep architecture for semantic matching with multiple positional
  sentence representations}.
In: \bbtitle{AAAI}
(\byear{2016})
\end{bchapter}
\endbibitem

\bibitem{Yin2009DetectionOH}
\begin{bchapter}
\bauthor{\bsnm{Yin}, \binits{D.}},
\bauthor{\bsnm{Xue}, \binits{Z.}},
\bauthor{\bsnm{Hong}, \binits{L.}},
\bauthor{\bsnm{Davison}, \binits{B.D.}},
\bauthor{\bsnm{Edwards}, \binits{L.}}:
\bctitle{Detection of harassment on web 2.0}.
(\byear{2009})
\end{bchapter}
\endbibitem

\bibitem{Sadiq2021AggressionDT}
\begin{barticle}
\bauthor{\bsnm{Sadiq}, \binits{S.}},
\bauthor{\bsnm{Mehmood}, \binits{A.}},
\bauthor{\bsnm{Ullah}, \binits{S.}},
\bauthor{\bsnm{Ahmad}, \binits{M.}},
\bauthor{\bsnm{Choi}, \binits{G.S.}},
\bauthor{\bsnm{On}, \binits{B.-W.}}:
\batitle{Aggression detection through deep neural model on twitter}.
\bjtitle{Future Gener. Comput. Syst.}
\bvolume{114},
\bfpage{120}--\blpage{129}
(\byear{2021})
\end{barticle}
\endbibitem

\bibitem{Devlin2019BERTPO}
\begin{bchapter}
\bauthor{\bsnm{Devlin}, \binits{J.}},
\bauthor{\bsnm{Chang}, \binits{M.-W.}},
\bauthor{\bsnm{Lee}, \binits{K.}},
\bauthor{\bsnm{Toutanova}, \binits{K.}}:
\bctitle{Bert: Pre-training of deep bidirectional transformers for language
  understanding}.
In: \bbtitle{NAACL}
(\byear{2019})
\end{bchapter}
\endbibitem

\end{thebibliography}
\bibliographystyle{spmpsci}


\end{document}